# Range-only Collaborative Localization for Ground Vehicles

Qin Shi, Xiaowei Cui, Sihao Zhao, Jian Wen, Mingquan Lu*, *Tsinghua University, China*
*Beijing National Research Center for Information Science and Technology, China

**BIOGRAPHY (IES)**

**Qin Shi** is a Ph.D. candidate in the Department of Electronic Engineering at Tsinghua University, China. He received his B.S. degree from Tsinghua University in 2015. His current research focuses on collaborative localization, sensor fusion and simultaneous localization and mapping.
**Xiaowei Cui** is an associate professor at the Department of Electronic Engineering, Tsinghua University, China. He is a member of the Expert Group of China BeiDou Navigation Satellite System. His research interests include robust GNSS signal processing, multipath mitigation techniques, and high-precision positioning.
**Sihao Zhao** is a lecturer at the Department of Electronic Engineering, Tsinghua University. He received the BS and PhD degrees from the Department of Electronic Engineering, Tsinghua University in 2005 and 2011, respectively.
**Jian Wen** is a Ph.D. candidate in the Department of Electronic Engineering at Tsinghua University, China. He received his B.Eng. (2015) in Communication Engineering at Beijing Jiaotong University, Beijing, China. His current research focuses on GNSS anti-spoofing and locating GNSS spoofers.
**Mingquan Lu** is a professor of the Department of Electronic Engineering, Tsinghua University, China. He is the director of Tsinghua Position, Navigation and Timing Center, and a member of the Expert Group of China BeiDou Navigation Satellite System. His current research interests include GNSS signal design and analysis, GNSS signal processing and receiver development, and GNSS system modeling and simulation.

## ABSTRACT

High-accuracy absolute localization for a team of vehicles is essential when accomplishing various kinds of tasks. As a promising approach, collaborative localization fuses the individual motion measurements and the inter-vehicle measurements to collaboratively estimate the states. In this paper, we focus on the range-only collaborative localization, which specifies the inter-vehicle measurements as inter-vehicle ranging measurements. We first investigate the observability properties of the system and derive that to achieve bounded localization errors, two vehicles are required to remain static like external infrastructures. Under the guide of the observability analysis, we then propose our range-only collaborative localization system which categorize the ground vehicles into two static vehicles and dynamic vehicles. The vehicles are connected utilizing a UWB network that is capable of both producing inter-vehicle ranging measurements and communication. Simulation results validate the observability analysis and demonstrate that collaborative localization is capable of achieving higher accuracy when utilizing the inter-vehicle measurements. Extensive experimental results are performed for a team of 3 and 5 vehicles. The real-world results illustrate that our proposed system enables accurate and real-time estimation of all vehicles' absolute poses.

## INTRODUCTION

Precise localization is undoubtedly a fundamental module for vehicle autonomy. A vehicle needs to know its absolute position and orientation, in indoor and outdoor environments, to achieve autonomous navigation and then perform high-level applications. Since the capability of a single agent is limited, recently we have seen the appearance of connected vehicles for a wide range of applications, such as disaster assessment [1], border surveillance [2], and planet exploration [3]. Consequently, the localization of a team of connected vehicles has gained significant interests in the research fields.

Using external infrastructures is a practical solution. Vehicles equipped with *exteroceptive* sensors are able to collect vehicle-infrastructure information to determine their states. Most of the existing approaches are first developed for the case of a single vehicle. For example, the well-known Global Navigation Satellite System (GNSS) [4], optical motion capture systems [5], and the ultra-wide band (UWB) localization systems [6] can estimate a vehicle's state with respect to the reference established by infrastructures. For a team of, say $N$, vehicles within the measuring range of infrastructures, the localization problem is then addressed by independently solving $N$ state estimation problems. In the context of simultaneous localization and mapping (SLAM), a vehicle can determine its pose with respect to the *virtual* infrastructures, say environment landmarks using *exteroceptive* information gathered by onboard cameras and/or Lidars [7]. A map representing the environment is then established, and a team of vehicles can then individually determine their pose by finding the correspondence with this same map. A more precise localization scheme is to optimally fuse the measurements from *proprioceptive* sensors that provide motion information, such as inertial measurement unit (IMU) and wheel encoders, with *exteroceptive* sensors. For example, the integrated GPS/IMU systems [8] and visual-inertial systems (VINS) [9] have proven to provide a significant improvement by bridging the gap between losses of *exteroceptive* information. However, in all cases, each vehicle estimates its state using only individual information, the inter-vehicle information is not combined.

A much preferable approach to preserving the vehicle team autonomy is to collaboratively localize the vehicles. In this case, the inter-vehicle measurements are collected by onboard *exteroceptive* sensors and shared among the vehicles to fuse with their motion information collected by their individual *proprioceptive* sensors. By information sharing, collaborative localization is anticipated to provide better performance, in terms of localization accuracy, scalability, robustness and sensor coverage. The inter-vehicle measurements can be categorized into the relative pose, the bearing and the range between the vehicles. Most existing approaches focus on the collaborative localization using the relative pose measurements [10-11]. Usually, a vision-based approach is used to detect the relative pose of neighbors by recognizing the unique localization image patterns attached to neighboring vehicles [12]. However, the natural weakness of visual methods, such as a limited field of view and motion blur prevents practical implementation of relative pose based collaborative localization. Recently, we have seen a growing trend of utilizing UWB technology as a mean of measuring the inter-vehicle range [13-14]. The fine temporal resolution and robustness to multipath enable reliable and precise direct inter-agent ranging measurements [15]. Together, this suggests that a setup with inter-vehicle ranging sensors and *proprioceptive* sensors is suitable for a collaborative localization system. However, unlike the relative pose based collaborative localization which has been well-studied and developed, range-only collaborative localization still remains challenging. Its basic theories in observability analysis are currently not thoroughly studied, and no off-the-shelf products can be found.

Observability analysis is to study the observability property of a system. If the system is observable, it contains all the necessary information to estimate its states with a bounded error. An early result of the observability analysis for a collaborative localization system was presented in [10]. They used a linear approximation method to analyze a multi-vehicle system, in which relative pose measurements and motion measurements are provided. The results show that the system is unobservable unless one of the vehicles have absolute position and orientation information. [16] also showed that the relative-pose based collaborative localization system has unobservable subspace, thus leading to inconsistency of the standard EKF estimator. Siegwart et al. [17] studied the observability property for a pair of two vehicles by considering the system nonlinearities. They separately investigate the range and bearing measurements and showed that with either type of measurement, the system is not fully observable. A deeper observability analysis for range-only collaborative localization system was studied in [18]. They proved that if 5 inter-vehicle range measurements are provided, the system is then observable, and the vehicle-to-vehicle relative pose can be estimated. However, they only studied the observability properties of two robots, and the collaborative localization problem was resolved with respect to a relative reference, i.e., the pose of the second vehicle was estimated with respect to the first vehicle. An inspiring work focusing on the observability analysis of bearing-only collaborative localization system is found in [19]. They derived the maximum rank of the observability matrix without global information and show that to achieve full observability, all nodes must connect to at least two external infrastructures with known location. However, for range-only collaborative localization systems, the observability properties with respect to a global absolute reference is currently not well studied, and the conditions under which the system is observable are not determined.

Despite the challenges in theoretical analysis, the challenges also arise from technical engineering. The first one lies in the inter-vehicle ranging implementation. By measuring the time-of-arrival (TOA) of UWB signals, the range between two vehicles can be obtained. Conventional inter-vehicle ranging method requires two-way message exchange due to the unavailability of clock synchronization in a pair of vehicles [20]. However, this is time-consuming for a team of $N$ vehicles, in which case the requirement of $N(N-1)/2$ times of two-way ranging leads to a significant latency of state estimation. The second challenge is that collaboration requires each vehicle is equipped with a communication module for sharing distributed measurements. The high sampling rate of the sensors may cause substantial communication overhead, thus leading to a latency of measurements sharing. Another challenge lies in the rigorous initialization. Typically, the system is launched with each vehicle at an unknown initial state. To directly fuse the inter-vehicle ranging measurements with motion measurements, a common global absolute reference and the initial poses of each vehicle in this reference are required.

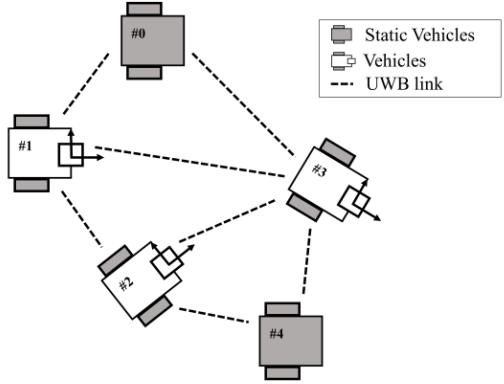

Fig.1. The UWB connected vehicles in our proposed range-only collaborative localization. Each vehicle is equipped with a UWB transceiver and a wheel encoder. Two vehicles with unknown positions are selected to remain static. Our aim is to perform collaborative localization by fusing the UWB ranging measurements with motion measurements.

In this paper, we propose a range-only collaborative localization system for a team of UWB connected ground vehicles, with the assistance of wheel encoders. We first give the observability analysis of a range-only collaborative localization system. To be fully observable of the vehicle states with respect to a global absolute reference, we derive a lemma that at least two vehicles are needed to be static to serve as infrastructures. Utilizing this lemma, we then present a complete UWB based range-only collaborative localization system for a team of ground vehicles. In our system, each ground vehicle is equipped with a UWB transceiver and a wheel encoder that provides instant linear velocity and turn rate measurements. Among them, two vehicles with unknown positions are selected to serve as external infrastructures as illustrated in Fig. 1. The main contributions of this paper are summarized as follows.

1) We analyze the observability properties of the error-state system model, and show that to achieve full observability of a team of connected ground vehicles, range-only collaborative localization requires two vehicles remaining static;
2) Based on the observability analysis, we propose our UWB-based range-only collaborative localization system, with robust initialization procedure to bootstrap the system with each vehicle at unknown initial states;
3) We implement our system using a customized UWB ranging and communication network, achieving real-world usability with a centimeter accuracy.

**SYSTEM MODELS**

Consider a team of $N(N \geq 3)$ ground vehicles moving in a horizontal plane performing collaborative localization. Each vehicle $R_i$ outputs its instant linear velocity $v_i$ and angular speed $\omega_i$. Meanwhile, each pair of vehicles $(R_i, R_j)$ estimates their ranges $d_{ij}$. Our goal is to fuse these measurements to collaboratively estimate the states of all vehicles $\mathbf{x} = [\mathbf{x}_1^T, \dots, \mathbf{x}_N^T]^T$, where $\mathbf{x}_i$ is the state of vehicle $R_i$. It is defined as $\mathbf{x}_i = [\mathbf{p}_i, \theta_i]^T \in \mathbb{R}^3$, where $\mathbf{p}_i = [x_i, y_i]$ is the planar position and $\theta_i$ is the robot heading with respect to an absolute reference.

The linear velocity and angular speed measurements are affected by additive noise:
$$v_i = v_{t,i} + n_{v,i}$$
$$\omega_i = \omega_{t,i} + n_{\omega,i}$$
where $v_{t,i}$ and $\omega_{t,i}$ are true values and the additive noise are assumed to be Gaussian white noise, $n_{v,i} \sim N(0, \sigma_{v,i})$, $n_{\omega,i} \sim N(0, \sigma_{\omega,i})$. We can then write the kinematic equations for $R_i$ as

$$\dot{\mathbf{x}}_i = f(\mathbf{x}_i, \mathbf{u}_i) = \begin{bmatrix} \cos(\theta_i) v_i \\ \sin(\theta_i) v_i \\ \omega_i \end{bmatrix} \quad (1)$$

where $\mathbf{u}_i = [v_i, 0, \omega_i]^T$ is the input vector.

In the proposed system, it is assumed that each vehicle moves in the sensory and communication range of its neighbors, i.e., the vehicles are connected all the time and the range measurements to its neighbors are available. The range between $R_i$ and $R_j$ is modeled as:

$$d_{ij} = h(\mathbf{x_i}, \mathbf{x_j}) + n_{ij} = \|\mathbf{p}_i - \mathbf{p}_j\| + n_{ij} \tag{2}$$

Where $n_{ij}$ is the measurement noise which is assumed to be a Gaussian noise, $n_{ij} \sim N(0, \sigma_{ij}^d)$.

## OBSERVABILITY ANALYSIS

A system is observable if the results of the measurements of the output provides enough information for determining its states. This means that given necessary and sufficient measurements, the states of the system can be estimated with bounded error. For this reason, it is important to study the observability property of the system. For nonlinear systems, a much preferable approach is to employ nonlinear observability tools. In this section, we apply the observability rank condition based on Lie derivatives [21] to study the observability of our range-only collaborative localization system and obtain the conditions under which the system is fully observable.

### Nonlinear Observability

For the sake of clarity, we first outline the nonlinear observability test employed in this paper. Consider a nonlinear system $\Sigma$ with the state-space representation as follows:

$$\Sigma: \begin{matrix} \dot{\mathbf{x}} = f(\mathbf{x}, \mathbf{u}) \\ \mathbf{y} = h(\mathbf{x}) \end{matrix} \tag{3}$$

where $\mathbf{x} \in \mathbb{R}^n$ is the state vector, $\mathbf{u} = [u_1, \ldots, u_l]^T \in \mathbb{R}^l$ control input vector, and $\mathbf{y} = [y_1, \ldots, y_m]^T \in \mathbb{R}^m$ is the measurement vector, with $y_k = h_k(\mathbf{x}), k = 1, \ldots, m$.

We consider the special case of our system where the process function $f$ is input-linear, which can be separated into a summation of independent functions, each one corresponds to a component of the control input vector:

$$\dot{\mathbf{x}} = f(\mathbf{x}, \mathbf{u}) = f_1(\mathbf{x})u_1 + \cdots + f_l(\mathbf{x})u_l = f_{v_1}(\mathbf{x})v_1 + f_{\omega_1}(\mathbf{x})\omega_1 + \cdots + f_{v_N}(\mathbf{x})v_N + f_{\omega_N}(\mathbf{x})\omega_N \tag{4}$$

where $l = 2N$.

The zeroth-order Lie derivative of any (scalar) function is the function itself, i.e., $\mathcal{L}^0 h_k(\mathbf{x}) = h_k(\mathbf{x})$. The first-order Lie derivative of function $h_k(\mathbf{x})$ with respect to $f_i$ is defined as:

$$\mathcal{L}_{f_i}^1 h_k(\mathbf{x}) = \frac{\partial h_k(\mathbf{x})}{\partial x_1} f_{i1}(\mathbf{x}) + \cdots + \frac{\partial h_k(\mathbf{x})}{\partial x_n} f_{in}(\mathbf{x}) = \nabla h_k(\mathbf{x}) \cdot f_i(\mathbf{x}) \tag{5}$$

where $f_i(\mathbf{x}) = [f_{i1}(\mathbf{x}), \ldots, f_{in}(\mathbf{x})]^T$, $n = 3N$, $\nabla$ denotes the gradient operator, and $\cdot$ represents the vector inner product. Considering that $\mathcal{L}_{f_i}^1 h_k(\mathbf{x})$ is a scalar function itself, the second-order Lie derivative of $h_k(\mathbf{x})$ with respect to $f_i$ can be computed recursively as:

$$\mathcal{L}_{f_i}^2 h_k(\mathbf{x}) = \mathcal{L}_{f_i}^1 \left( \mathcal{L}_{f_i}^1 h_k(\mathbf{x}) \right) = \nabla \mathcal{L}_{f_i}^1 h_k(\mathbf{x}) \cdot f_i(\mathbf{x}) \tag{6}$$

Higher order Lie derivatives can be computed similarly. The mixed Lie derivatives is defined, e.g. the second-order Lie derivative of $h_k(\mathbf{x})$ with respect to $f_i$ and $f_j$, given its first derivative with respect to $f_i$, is:

$$\mathcal{L}_{f_i f_j}^2 h_k(\mathbf{x}) = \mathcal{L}_{f_j}^1 \left( \mathcal{L}_{f_i}^1 h_k(\mathbf{x}) \right) = \nabla \mathcal{L}_{f_i}^1 h_k(\mathbf{x}) \cdot f_j(\mathbf{x}) \tag{7}$$

The observability matrix is defined as the matrix whose rows are the gradients of Lie derivatives:

$$\mathcal{O} \triangleq \left\{ \nabla \mathcal{L}_{f_i \ldots f_j}^\ell h_k(\mathbf{x}) \,\middle|\, i, j = 0, \ldots, l; k = 1, \ldots, m; \ell \in \mathbb{N} \right\} \tag{8}$$

*Proposition 1 (Observability rank condition):* The nonlinear system defined in (3) is locally weakly observable if its observability matrix (8) has full rank.

### Observability with only dynamic vehicles

We first consider a two-vehicle case (e.g., $R_i$ and $R_j$). We rearrange the nonlinear kinematic equations in the following form for convenient Lie derivatives computation:

$$\dot{\mathbf{x}} = \begin{bmatrix} \dot{\mathbf{x}}_i \\ \dot{\mathbf{x}}_j \end{bmatrix} = \boldsymbol{f}_{v_i} v_i + \boldsymbol{f}_{\omega_i} \omega_i + \boldsymbol{f}_{v_j} v_j + \boldsymbol{f}_{\omega_j} \omega_j = \begin{bmatrix} \cos(\theta_i) \\ \sin(\theta_i) \\ 0 \\ 0 \\ 0 \\ 0 \end{bmatrix} v_i + \begin{bmatrix} 0 \\ 0 \\ 1 \\ 0 \\ 0 \\ 0 \end{bmatrix} \omega_i + \begin{bmatrix} 0 \\ 0 \\ 0 \\ \cos(\theta_j) \\ \sin(\theta_j) \\ 0 \end{bmatrix} v_j + \begin{bmatrix} 0 \\ 0 \\ 0 \\ 0 \\ 0 \\ 1 \end{bmatrix} \omega_j$$

Furthermore, to preserve the clarity of presentation, we choose the measurement function to be the squared range between two vehicles divided by two, i.e.,

$$h(\mathbf{x}) = \frac{d_{ij}^2}{2} = \frac{1}{2}(\mathbf{p}_i - \mathbf{p}_j)^T(\mathbf{p}_i - \mathbf{p}_j)$$

Note that since d is positive, there is a bijection between d and $d^2/2$, and they provide the same information on the system observability. We then give the corresponding Lie derivatives and their gradients to derive the observability matrix of the two-vehicle system.

1) *Zeroth-order Lie derivative:*

$$\mathcal{L}^0 h = h = \frac{1}{2}(\mathbf{p}_i - \mathbf{p}_j)^T(\mathbf{p}_i - \mathbf{p}_j)$$

with gradient:

$$\nabla \mathcal{L}^0 h = [\Delta x_{ij} \quad \Delta y_{ij} \quad 0 \quad -\Delta x_{ij} \quad -\Delta y_{ij} \quad 0]$$

where $\Delta x_{ij} = x_i - x_j$ and $\Delta y_{ij} = y_i - y_j$.

2) *First-order Lie derivatives:*

$$\mathcal{L}^1_{\boldsymbol{f}_{v_i}} h = \nabla \mathcal{L}^0 h \cdot \boldsymbol{f}_{v_i} = \cos(\theta_i) \Delta x_{ij} + \sin(\theta_i) \Delta y_{ij}$$
$$\mathcal{L}^1_{\boldsymbol{f}_{\omega_i}} h = \nabla \mathcal{L}^0 h \cdot \boldsymbol{f}_{\omega_i} = 0$$
$$\mathcal{L}^1_{\boldsymbol{f}_{v_j}} h = \nabla \mathcal{L}^0 h \cdot \boldsymbol{f}_{v_j} = -\cos(\theta_j) \Delta x_{ij} - \sin(\theta_j) \Delta y_{ij}$$
$$\mathcal{L}^1_{\boldsymbol{f}_{\omega_j}} h = \nabla \mathcal{L}^0 h \cdot \boldsymbol{f}_{\omega_j} = 0$$

with gradients:

$$\nabla \mathcal{L}^1_{\boldsymbol{f}_{v_i}} h = [\cos(\theta_i) \quad \sin(\theta_i) \quad -D_i^- \quad -\cos(\theta_i) \quad -\sin(\theta_i) \quad 0]$$
$$\nabla \mathcal{L}^1_{\boldsymbol{f}_{v_j}} h = [-\cos(\theta_j) \quad -\sin(\theta_j) \quad 0 \quad \cos(\theta_j) \quad \sin(\theta_j) \quad D_j^-]$$

where $D_i^- = \sin(\theta_i) \Delta x_{ij} - \cos(\theta_i) \Delta y_{ij}$, and $D_j^- = \sin(\theta_j) \Delta x_{ij} - \cos(\theta_j) \Delta y_{ij}$. We have omitted the gradients of $\mathcal{L}^1_{\boldsymbol{f}_{\omega_i}} h$ and $\mathcal{L}^1_{\boldsymbol{f}_{\omega_j}} h$ since they are zero vectors.

3) *Second-order Lie derivatives:*

$$\mathcal{L}^2_{\boldsymbol{f}_{v_i}\boldsymbol{f}_{v_i}} h = \nabla \mathcal{L}^1_{\boldsymbol{f}_{v_i}} h \cdot \boldsymbol{f}_{v_i} = 1$$
$$\mathcal{L}^2_{\boldsymbol{f}_{v_i}\boldsymbol{f}_{v_j}} h = \nabla \mathcal{L}^1_{\boldsymbol{f}_{v_i}} h \cdot \boldsymbol{f}_{v_j} = -\cos(\theta_i - \theta_j)$$
$$\mathcal{L}^2_{\boldsymbol{f}_{v_j}\boldsymbol{f}_{v_i}} h = \nabla \mathcal{L}^1_{\boldsymbol{f}_{v_j}} h \cdot \boldsymbol{f}_{v_i} = -\cos(\theta_i - \theta_j)$$
$$\mathcal{L}^2_{\boldsymbol{f}_{v_j}\boldsymbol{f}_{v_j}} h = \nabla \mathcal{L}^1_{\boldsymbol{f}_{v_j}} h \cdot \boldsymbol{f}_{v_j} = 1$$
$$\mathcal{L}^2_{\boldsymbol{f}_{v_i}\boldsymbol{f}_{\omega_i}} h = \nabla \mathcal{L}^1_{\boldsymbol{f}_{v_i}} h \cdot \boldsymbol{f}_{\omega_i} = -D_i^-$$
$$\mathcal{L}^2_{\boldsymbol{f}_{v_i}\boldsymbol{f}_{\omega_j}} h = \nabla \mathcal{L}^1_{\boldsymbol{f}_{v_i}} h \cdot \boldsymbol{f}_{\omega_j} = 0$$
$$\mathcal{L}^2_{\boldsymbol{f}_{v_j}\boldsymbol{f}_{\omega_i}} h = \nabla \mathcal{L}^1_{\boldsymbol{f}_{v_j}} h \cdot \boldsymbol{f}_{\omega_i} = 0$$
$$\mathcal{L}^2_{\boldsymbol{f}_{v_j}\boldsymbol{f}_{\omega_j}} h = \nabla \mathcal{L}^1_{\boldsymbol{f}_{v_j}} h \cdot \boldsymbol{f}_{\omega_j} = D_j^-$$

with gradients:

$$\nabla \mathcal{L}^2_{\boldsymbol{f}_{v_i}\boldsymbol{f}_{v_j}} h = [0 \quad 0 \quad \sin(\theta_i - \theta_j) \quad 0 \quad 0 \quad -\sin(\theta_i - \theta_j)]$$
$$\nabla \mathcal{L}^2_{\boldsymbol{f}_{v_i}\boldsymbol{f}_{\omega_i}} h = [-\sin(\theta_i) \quad \cos(\theta_i) \quad -D_i^+ \quad \sin(\theta_i) \quad -\cos(\theta_i) \quad 0]$$
$$\nabla \mathcal{L}^2_{\boldsymbol{f}_{v_j}\boldsymbol{f}_{\omega_j}} h = [\sin(\theta_j) \quad -\cos(\theta_j) \quad 0 \quad -\sin(\theta_j) \quad \cos(\theta_j) \quad D_j^+]$$

where $D_i^+ = \cos(\theta_i)\Delta x_{ij} + \sin(\theta_i)\Delta y_{ij}$ and $D_j^+ = \cos(\theta_j)\Delta x_{ij} + \sin(\theta_j)\Delta y_{ij}$. Gradient of $\mathcal{L}^2_{f_{v_j}f_{v_i}}h$ is linearly dependent on $\nabla\mathcal{L}^2_{f_{v_i}f_{v_j}}h$ and gradients of $\mathcal{L}^2_{f_{v_i}f_{v_i}}h, \mathcal{L}^2_{f_{v_j}f_{v_j}}h, \mathcal{L}^2_{f_{v_i}f_{\omega_j}}h$ and $\mathcal{L}^2_{f_{v_j}f_{\omega_i}}h$ are $\mathbf{0}$, thus are omitted. Such gradients are also omitted when computing higher-order Lie derivatives.

4) *Third-order Lie derivatives:*

$$\mathcal{L}^3_{f_{v_i}f_{v_j}f_{\omega_i}}h = \nabla\mathcal{L}^2_{f_{v_i}f_{v_j}}h \cdot f_{\omega_i} = \sin(\theta_i - \theta_j)$$
$$\mathcal{L}^3_{f_{v_i}f_{\omega_i}f_{\omega_i}}h = \nabla\mathcal{L}^2_{f_{v_i}f_{\omega_i}}h \cdot f_{\omega_i} = -D_i^+$$
$$\mathcal{L}^3_{f_{v_j}f_{\omega_j}f_{\omega_j}}h = \nabla\mathcal{L}^2_{f_{v_j}f_{\omega_j}}h \cdot f_{\omega_j} = D_j^+$$

with gradients:

$$\nabla\mathcal{L}^3_{f_{v_i}f_{v_j}f_{\omega_i}}h = \begin{bmatrix} 0 & 0 & \cos(\theta_i - \theta_j) & 0 & 0 & -\cos(\theta_i - \theta_j) \end{bmatrix}$$
$$\nabla\mathcal{L}^3_{f_{v_i}f_{\omega_i}f_{\omega_i}}h = \begin{bmatrix} -\cos(\theta_i) & -\sin(\theta_i) & D_i^- & \cos(\theta_i) & \sin(\theta_i) & 0 \end{bmatrix}$$
$$\nabla\mathcal{L}^3_{f_{v_j}f_{\omega_j}f_{\omega_j}} = \begin{bmatrix} \cos(\theta_j) & \sin(\theta_j) & 0 & -\cos(\theta_j) & -\sin(\theta_j) & -D_j^- \end{bmatrix}$$

Clearly, $\nabla\mathcal{L}^3_{f_{v_i}f_{\omega_i}f_{\omega_i}}h = -\nabla\mathcal{L}^1_{f_{v_i}}h$ and $\nabla\mathcal{L}^3_{f_{v_j}f_{\omega_j}f_{\omega_j}} = -\nabla\mathcal{L}^1_{f_{v_j}}h$. Higher-order Lie derivatives are linearly dependent on third-order and lower-order Lie derivatives. Therefore, we stack together all the previously computed linearly independent gradients of Lie derivatives to form the observability matrix of the two-vehicle system:

$$\mathcal{O}_{ij} = \begin{bmatrix} \Delta x_{ij} & \Delta y_{ij} & 0 & -\Delta x_{ij} & -\Delta y_{ij} & 0 \\ \cos(\theta_i) & \sin(\theta_i) & -D_i^- & -\cos(\theta_i) & -\sin(\theta_i) & 0 \\ -\cos(\theta_j) & -\sin(\theta_j) & 0 & \cos(\theta_j) & \sin(\theta_j) & D_j^- \\ 0 & 0 & \sin(\theta_i - \theta_j) & 0 & 0 & -\sin(\theta_i - \theta_j) \\ -\sin(\theta_i) & \cos(\theta_i) & -D_i^+ & \sin(\theta_i) & -\cos(\theta_i) & 0 \\ \sin(\theta_j) & -\cos(\theta_j) & 0 & -\sin(\theta_j) & \cos(\theta_j) & D_j^+ \\ 0 & 0 & \cos(\theta_i - \theta_j) & 0 & 0 & -\cos(\theta_i - \theta_j) \end{bmatrix}$$

*Proposition 2 (Full rank factorizations [22]):* Let $\mathbb{R}_r^{m\times n}$ *be the collection of the matrices of rank r in* $\mathbb{R}^{m\times n}$, *every matrix* $A \in \mathbb{R}_r^{m\times n}$ *with* $r > 0$ *then have a full rank factorization of* $A = FG$, *where* $F \in \mathbb{R}_r^{m\times r}$ *and* $G \in \mathbb{R}_r^{r\times n}$. *One algorithm to compute the full rank factorization of* A *is by applying a finite sequence of elementary row operations to transform matrix* A *into the row reduced echelon form, RREF(A). The matrix F can be constructed from the columns of A that correspond to the columns with the leading ones in RREF(A), and the matrix G can be constructed from the nonzero rows of RREF(A).*

*Lemma 1:* The rank of the observability matrix $\mathcal{O}_{ij}$, of the range-only collaborative localization system in the two-vehicle case, is equal to 3.
*Proof:* By applying a finite sequence of elementary row operations, we can produce the unique row reduced echelon form of $\mathcal{O}_{ij}$, RREF($\mathcal{O}_{ij}$) in the form as:

$$\text{RREF}(\mathcal{O}_{ij}) = \begin{bmatrix} \mathbf{I}_3 & G_{ij} \\ \mathbf{0}_{4\times 3} & \mathbf{0}_{4\times 3} \end{bmatrix}$$

where

$$G_{ij} = \begin{bmatrix} -1 & 0 & \Delta y \\ 0 & -1 & -\Delta x \\ 0 & 0 & -1 \end{bmatrix}$$

Clearly, $G_{ij}$ has three linearly independent rows. From Proposition 2, we can derive that $\text{rank}(\mathcal{O}_{ij}) = \text{rank}(\text{RREF}(\mathcal{O}_{ij})) = 3$.

Wo now study the general case of n moving vehicles. Proceeding similarly to the two-vehicle case, we can write the row reduced echelon form of corresponding observability matrix $\mathcal{O}^n$, RREF($\mathcal{O}^n$) as:

$$\text{RREF}(\mathcal{O}^n) = \{\text{RREF}(\mathcal{O}_{ij}^2)\}, i, j = 1, \dots, n$$

where $\text{RREF}(\mathcal{O}_{ij}^2) = [\mathbf{0}_{3\times 3(i-1)} \quad \mathbf{I}_3 \quad \mathbf{0}_{3\times(3(j-1)-3i)} \quad G_{ij} \quad \mathbf{0}_{3\times 3(n-j)}]$. Note that we rewrite the RREF($\mathcal{O}_{ij}$) as the collection its linearly independent rows since they span the same observable space as *Proposition 2* states.

We first consider the observability properties of three connected vehicles.
*Lemma 2:* The rank of the observability matrix of three connected moving vehicles is equal to 6.

*Proof:* Since three vehicles are fully connected, we can write its corresponding RREF($\mathcal{O}^3$) as:
$$\text{RREF}(\mathcal{O}^3) = \begin{bmatrix} \text{RREF}(\mathcal{O}_{13}^3) \\ \text{RREF}(\mathcal{O}_{23}^3) \\ \text{RREF}(\mathcal{O}_{12}^3) \end{bmatrix} = \begin{bmatrix} I_3 & 0 & G_{13} \\ 0 & I_3 & G_{23} \\ I_3 & G_{12} & 0 \end{bmatrix}$$

By applying elementary matrix $R = \begin{bmatrix} I_3 & 0 & 0 \\ 0 & I_3 & 0 \\ -I_3 & -G_{12} & I_3 \end{bmatrix}$ to RREF($\mathcal{O}^3$), we have:

$$R \cdot \text{RREF}(\mathcal{O}^3) = \begin{bmatrix} I_3 & 0 & G_{13} \\ 0 & I_3 & G_{23} \\ 0 & 0 & -G_{13} - G_{12}G_{23} \end{bmatrix} = \begin{bmatrix} I_3 & 0 & G_{13} \\ 0 & I_3 & G_{23} \\ 0 & 0 & 0 \end{bmatrix}$$

Therefore, rank($\mathcal{O}^3$) = 6 as *Proposition 2* implies.

Our next goal is to extend the above lemma to n vehicle case.

*Lemma 3: The rank of the observability matrix of the general n connected vehicles, is equal to* $3(n-1)$.
*Proof:* We first give the row reduced echelon form of the associated observability matrix in a recursive way. Base case with $n = 2$ and $n = 3$ have been showed in Lemma 1 and Lemma 2:
*for n = 2*, RREF($\mathcal{O}^2$) = [RREF($\mathcal{O}_{12}^2$)] = $[I_3 \quad G_{12}]$;
*for n = 3*, RREF($\mathcal{O}^3$) = $\begin{bmatrix} \text{RREF}(\mathcal{O}_{13}^3) \\ \text{RREF}(\mathcal{O}_{23}^3) \end{bmatrix} = \begin{bmatrix} I_3 & 0 & G_{13} \\ 0 & I_3 & G_{23} \end{bmatrix}$

Suppose when $n = k$, RREF($\mathcal{O}^k$) has the form of
$$\text{RREF}(\mathcal{O}^k) = \begin{bmatrix} \text{RREF}(\mathcal{O}_{1k}^k) \\ \text{RREF}(\mathcal{O}_{2k}^k) \\ \vdots \\ \text{RREF}(\mathcal{O}_{k-1,k}^k) \end{bmatrix} = \begin{bmatrix} I_3 & 0 & 0 & \cdots & G_{1k} \\ 0 & I_3 & 0 & \cdots & G_{2k} \\ \vdots & \vdots & \ddots & & \vdots \\ 0 & 0 & \cdots & I_3 & G_{k-1,k} \end{bmatrix}$$

We can construct $\mathcal{O}^{k+1}$ for $n = k+1$ as follows:

$$\mathcal{O}^{k+1} = \begin{bmatrix} \text{RREF}(\mathcal{O}_{1,k+1}^{k+1}) \\ \text{RREF}(\mathcal{O}_{2,k+1}^{k+1}) \\ \vdots \\ \text{RREF}(\mathcal{O}_{k,k+1}^{k+1}) \\ \text{RREF}(\mathcal{O}^k) \quad \mathbf{0}_{3k \times 3} \end{bmatrix} = \begin{bmatrix} I_3 & 0 & 0 & \cdots & 0 & G_{1,k+1} \\ 0 & I_3 & 0 & \cdots & 0 & G_{2,k+1} \\ \vdots & \vdots & \ddots & & 0 & \vdots \\ 0 & 0 & \cdots & 0 & I_3 & G_{k,k+1} \\ I_3 & 0 & 0 & \cdots & G_{1k} & 0 \\ 0 & I_3 & 0 & \cdots & G_{2k} & 0 \\ \vdots & \vdots & \ddots & & \vdots & 0 \\ 0 & 0 & \cdots & I_3 & G_{k-1,k} & 0 \end{bmatrix}$$

Since we have $G_{ij}G_{jk} = -G_{ik}$, it is straightforward to derive RREF($\mathcal{O}^{k+1}$) as following:
$$\text{RREF}(\mathcal{O}^{k+1}) = \begin{bmatrix} \text{RREF}(\mathcal{O}_{1,k+1}^{k+1}) \\ \text{RREF}(\mathcal{O}_{2,k+1}^{k+1}) \\ \vdots \\ \text{RREF}(\mathcal{O}_{k,k+1}^{k+1}) \\ \mathbf{0}_{3k \times 3(k+1)} \end{bmatrix}$$

We then rewrite RREF($\mathcal{O}^{k+1}$) as follows since they span the same space:
$$\text{RREF}(\mathcal{O}^{k+1}) = \begin{bmatrix} \text{RREF}(\mathcal{O}_{1,k+1}^{k+1}) \\ \text{RREF}(\mathcal{O}_{2,k+1}^{k+1}) \\ \vdots \\ \text{RREF}(\mathcal{O}_{k,k+1}^{k+1}) \end{bmatrix}$$

To this end, we can conclude that for the observability matrix $\mathcal{O}^n$ of the n connected vehicles, its row reduced echelon form is given as:

$$\mathrm{RREF}(\mathcal{O}^n) = \begin{bmatrix} \mathrm{RREF}(\mathcal{O}_{1n}^n) \\ \mathrm{RREF}(\mathcal{O}_{2n}^n) \\ \vdots \\ \mathrm{RREF}(\mathcal{O}_{n-1,n}^n) \end{bmatrix} = \begin{bmatrix} \mathbf{I}_3 & \mathbf{0} & \mathbf{0} & \cdots & G_{1n} \\ \mathbf{0} & \mathbf{I}_3 & \mathbf{0} & \cdots & G_{2n} \\ \vdots & \vdots & & \ddots & \vdots \\ \mathbf{0} & \mathbf{0} & \cdots & \mathbf{I}_3 & G_{n-1,n} \end{bmatrix}$$

*Clearly,* $\mathrm{rank}(\mathcal{O}^n) = 3(n-1)$.

**Observability with static vehicles**

We have shown that the range-only collaborative localization system consisting of only dynamic vehicles, is not fully observable with respect to a global reference. To be fully observable, a common choice is to preinstall some static external infrastructures, e.g., UWB anchors, with known positions to provide information of the global reference. Instead, in this paper, we purposely control some of the vehicles to be static and serve as anchors, thus endowing system with the characterization of efficiency and easy deployment. In this section, we derive the number of vehicles need to be static to achieve full observability of our proposed range-only collaborative localization system.

We first derive the conditions of full observability for a single dynamic vehicle $R_i$ with unknown state $\mathbf{x}_i$.
Considering a static vehicle $R_k$ with known position $\mathbf{p}_k = [x_k, y_k]$, the corresponding nonlinear kinematic equations is then given as:

$$\dot{\mathbf{x}} = \dot{\mathbf{x}}_i = \boldsymbol{f}_{v_i} v_i + \boldsymbol{f}_{\omega_i} \omega_i = \begin{bmatrix} \cos(\theta_i) \\ \sin(\theta_i) \\ 0 \end{bmatrix} v_i + \begin{bmatrix} 0 \\ 0 \\ 1 \end{bmatrix} \omega_i$$

Similarly, let measurement function be the squared range between the two vehicles divided by two, i.e.,

$$h(\mathbf{x}) = \frac{d_{ik}^2}{2} = \frac{1}{2}(\mathbf{p}_i - \mathbf{p}_k)^T(\mathbf{p}_i - \mathbf{p}_k)$$

Proceeding similarly to previous subsection, the corresponding observability matrix is derived as:

$$\mathcal{O}_{ik} = \begin{bmatrix} \Delta x_{ik} & \Delta y_{ik} & 0 \\ \cos(\theta_i) & \sin(\theta_i) & -D_i^- \\ -\sin(\theta_i) & \cos(\theta_i) & -D_i^+ \end{bmatrix}$$

*Lemma 4: The rank of the observability matrix $\mathcal{O}_{ik}$, in the case of a dynamic vehicle and a static vehicle, is equal to 2.*
*Proof:* By applying elementary row operations to $\mathcal{O}_{ik}$, we give $\mathrm{RREF}(\mathcal{O}_{ik})$ in the form as:

$$\mathrm{RREF}(\mathcal{O}_{ik}) = \begin{bmatrix} G_{ik} \\ \mathbf{0}_{1\times 3} \end{bmatrix}$$

where

$$G_{ik} = \begin{bmatrix} 1 & 0 & \Delta y \\ 0 & 1 & -\Delta x \end{bmatrix}$$

Proposition 2 implies that $\mathrm{rank}(\mathcal{O}_{ik}) = \mathrm{rank}(G_{ik}) = 2$.

We then move to the case where two static vehicles are available.
*Lemma 5: For a dynamic vehicle, it is fully observable, i.e., the rank of the observability matrix is 3, if there exist two static vehicles serving as anchors with known positions.*
*Proof:* Without loss of generality, we let the static vehicles be $R_1$ and $R_2$. The corresponding $\mathrm{RREF}(\mathcal{O}_i)$ is then given as:

$$\mathrm{RREF}(\mathcal{O}_i) = \begin{bmatrix} G_{i1} \\ G_{i2} \end{bmatrix} = \begin{bmatrix} 1 & 0 & \Delta y_{i1} \\ 0 & 1 & -\Delta x_{i1} \\ 1 & 0 & \Delta y_{i2} \\ 0 & 1 & -\Delta x_{i2} \end{bmatrix}$$

By applying elementary row operations to $\mathrm{RREF}(\mathcal{O}_i)$, it is straightforward to have:

$$\mathrm{RREF}(\mathcal{O}_i) = \begin{bmatrix} \mathbf{I}_3 \\ \mathbf{0}_{1\times 3} \end{bmatrix}$$

Therefore, $\mathrm{rank}(\mathcal{O}_i) = 3$, the state of a single dynamic is fully observable if two static vehicles are available to provide additional range measurements.

The above lemma implies that measurements from two static vehicles will additionally provide three independent rows to the observability matrix. Therefore, for a team of n connected vehicles, if there are additional two static vehicles serving as anchors with known position, the range-only collaborative localization system is then fully observable, i.e., $\mathrm{rank}(\mathcal{O}) = 3n$.

## COLLABORATIVE LOCALIZATION IMPLEMENTATION

In this section, we illustrate the implementation details of our proposed range-only collaborative localization system. We start by restating the problem we are focusing on. For a team of $N(N \geq 3)$ ground vehicles first entering into an unknown environment, our goal is to estimate the combined states, $\mathbf{x} = [\mathbf{x}_1^T, \ldots, \mathbf{x}_N^T]^T$, providing that each vehicle $R_i$ can output its instant linear velocity $v_i$ and angular speed $\omega_i$ and each pair of vehicles $(R_i, R_j)$ can estimate their ranges $d_{ij}$.

Previous results of observability analysis imply that for N ground vehicles, the system is not fully observable if they are all dynamic ones. Therefore, we categorize the ground vehicles into dynamic vehicle set *DV* and static vehicle set *SV*. The number of the static vehicles is set to be 2. If the two static vehicles have their global position information, then the observability matrix of the remaining dynamic vehicles has rank of $3(N - 2 - 1) + 3 = 3(N - 2)$, which equals to the dimension of states to be estimated.

**System overview**
Our system mainly consists of two part: the UWB broadcast network that is capable of ranging and communication and a centralized collaborative localization estimator. The information communicated in the UWB network contains the 100Hz clock parameters of each vehicle with respect to its neighbors and the 20Hz motion measurements for each vehicle. We then collect all the information using a UWB sniffer by receiving the on-air UWB packets in a host server. The host server runs a centralized collaborative localization algorithm, whose processing pipeline is shown in Fig.2. The algorithm is mainly composed of three modules: 1) the preprocessing module that extracts information from the UWB network; 2) the motion-induced initializer that gives the global position of the two static vehicles and the initial states of the dynamic vehicles; 3) the estimator that recursively produce optimal states of the dynamic vehicles.

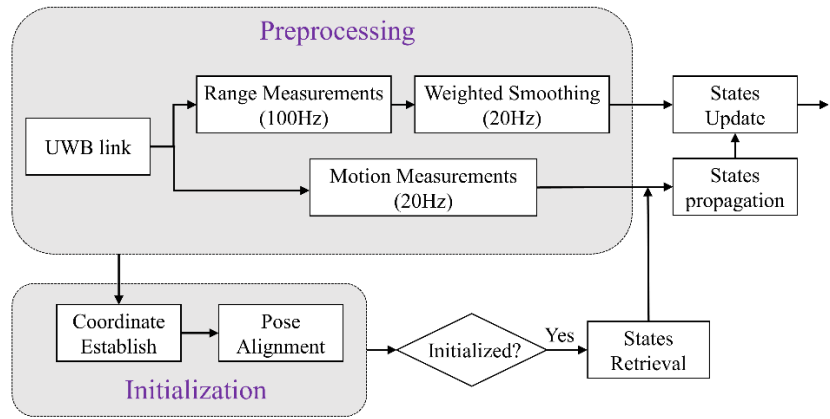

Fig.2. The pipeline of our proposed range-only collaborative localization system.

**UWB broadcast network**
We utilize a UWB network containing N transceivers for both ranging and communication. In detail, each UWB transceiver is scheduled to broadcast a packet using a time division multiple access (TDMA) scheme at a frequency of 100Hz. To achieve synchronization between vehicles, each vehicle hosts N-1 Kalman filters to maintain the clock parameters with the neighboring N-1 UWB transceivers. Then by collecting two-way packets, we can obtain all the vehicle-to-vehicle ranging measurements in one TDMA frame by extracting the time-of-flight (TOF) of the UWB signals. This enables high-frequency and real-time ranging measurements for the entirely connected vehicles and the clocks of each vehicle are then synchronized. Note that the information encoded in the broadcast packet contains not only the clock parameters utilized for ranging and clock synchronization but also the motion measurements from individual vehicle's wheel encoder. In this way, the measurements are shared among the connected vehicles using UWB communication links. We also note that the motion measurements are time synchronized using the UWB network time reference. The UWB sniffer then extracts the 20 Hz motion measurements and 100Hz pair-wise clock parameters from the UWB link. The 100Hz inter-vehicle ranging measurements are calculated using the extracted measurements and then smoothed using a weighted average method in the preprocessing module.

**Motion-induced initialization**

The motion-induced initializer starts with a coordinate establishing process to build a global reference which is bind to the two static vehicles. This is done by controlling all vehicles to remain static and then collecting the range measurements between all vehicles. The range measurements are utilized to form an adjacent matrix $D = [d_{ij}|i, j = 1, \ldots, N]$. A classical multidimensional scaling (MDS) method [23] is then used to find the rough position of each vehicle from the given adjacent matrix, with a particular choice of the coordinate system: the first static vehicle is fixed at the origin and the second static vehicle along the positive x-axis. Since the range measurements are corrupted with Gaussian noises (2), the positions of each vehicle are finally optimized by minimizing the following cost function:

$$L(\mathbf{p}) = \frac{1}{2}\sum_{i=1}^{N}\sum_{j=1,j\neq i}^{N}\left(d_{ij} - \|\mathbf{p}_i - \mathbf{p}_j\|\right)^2$$

where $\mathbf{p} = [\mathbf{p}_1^T, \ldots, \mathbf{p}_N^T]^T$. In this way, an optimal global reference (a right-hand coordinate) is established, with the two static vehicles global positions also determined with respect to this reference.

To derive the initial heading of each dynamic vehicle, the motion-induced initializer requires every dynamic vehicle performing linear motion for a while after coordinate established. If a vehicle is moving linearly, its trajectory then forms a line and its heading is equal to the angle between its trajectory line and positive x-axis. A virtual heading sensor is utilized. It uses the energy in the turn rate signal and the linear velocity signal to detect when the vehicle is moving linearly. The virtual heading sensor decides on linear motion for $R_i$ if:

$$E(\omega_i) = \frac{1}{M}\sum_{k=n}^{n+M-1}\|\omega_{i,k}\|^2 < \gamma_\omega$$

and

$$E(v_i) = \frac{1}{M}\sum_{k=n}^{n+M-1}\|v_{i,k}\|^2 > \gamma_v$$

where $E(\omega_i)$ is the turn rate energy, $E(v_i)$ is the linear velocity energy and $\omega_{i,k}$ and $v_{i,k}$ is the sampled measurement at step k. The threshold $\gamma$ is determined by the statistics of corresponding measurements, and is usually a scaling of $\sigma^2$. Once a linear motion is detected, the virtual heading sensor collects the range measurements with respect to the static two vehicles $R_{k1}$ and $R_{k2}$. The planar positions of the vehicle are then directly computed as:

$$x = d_{i,k1}\cos(\phi)$$
$$y = \pm d_{i,k2}\sin(\phi)$$

where $\phi = \frac{d_{i,k1}^2 + d_{i,k2}^2 - d_{k1,k2}^2}{2d_{i,k1}d_{k1,k2}}$. During initialization, we assume that the dynamic vehicles will not move through the x-axis, thus the y-ambiguity of the vehicle can be resolved by its initial y position computed in the coordinate establishing process. In this way, virtual heading sensor can determine the trajectory line of the moving vehicle and thus initialize its heading. To this end, the initial pose of every dynamic vehicle is determined and aligned to the global reference.

At this point, the initialization process is completed, and all the measurements can be resolved with respect to the established global reference and then fed into the estimator.

**Collaborative localization estimator**
The goal of the estimator is to estimate the combined states of the dynamic vehicles. We adopt the error-state Kalman filter (ESKF) [24] framework in our system.
**1) ESKF propagation**
The propagation process is dividually performed for each dynamic vehicle. For dynamic vehicle $R_i \in DV$, its states $\mathbf{x}_i$ are propagated using its wheel encoder measurements according to kinematic equation (2). For discrete-time implementation, we utilize the midpoint method for numerical integration, which use the velocity and turn rate measurements at time step k and $k + 1$ to linearly approximate the velocity and turn rate at the midpoint of the time interval from k to $k + 1$.

The ESKF propagate the covariance with respect to the error-state. Linearization of (2) yields the dynamics of error-state, which are then integrated into the discrete time system kinematics as follows:
$$\delta x_i \leftarrow f(x_i, \delta x_i, u_i, i_i) = F_{x_i}(x_i, u_i) \cdot \delta x_i + F_{i_i} \cdot i_i$$
where $\delta x_i$ is the error state vector, $\Delta t$ is the discrete time interval, $i_i$ is the perturbation impulses vector modeled as a white Gaussian process with covariance matrix

$$Q_i = \begin{bmatrix} (\cos(\theta_i)\sigma_{v,i}\Delta t)^2 & 0 & 0 \\ 0 & (\sin(\theta_i)\sigma_{v,i}\Delta t)^2 & 0 \\ 0 & 0 & (\sigma_{\omega,i}\Delta t)^2 \end{bmatrix}$$

$F_{x_i}$ and $F_{i_i}$ are the Jacobians of $f()$ with respect to the error state and the perturbation vector,

$$F_{x_i} = \frac{\partial f}{\partial \delta x_i}|_{x_i, u_i} = \begin{bmatrix} 1 & 0 & -\sin(\theta_i)v_i\Delta t \\ 0 & 1 & \cos(\theta_i)v_i\Delta t \\ 0 & 0 & 1 \end{bmatrix}$$

$$F_{i_i} = \frac{\partial f}{\partial i}|_{x, u_m} = I_3$$

The error-state kinematics of the entire system is then derived by stacking each vehicle's error-state kinematics:

$$\delta x \leftarrow \begin{bmatrix} F_{x_1} & \cdots & 0 \\ \vdots & \ddots & \vdots \\ 0 & \cdots & F_{x_n} \end{bmatrix} \begin{bmatrix} \delta x_1 \\ \vdots \\ \delta x_n \end{bmatrix} + \begin{bmatrix} I_3 & \cdots & 0 \\ \vdots & \ddots & \vdots \\ 0 & \cdots & I_3 \end{bmatrix} \begin{bmatrix} i_1 \\ \vdots \\ i_n \end{bmatrix} \triangleq F_x \cdot \delta x + F_i \cdot i$$

The covariance of the error state of the entire system is then propagated as:
$$P \leftarrow F_x P F_x^T + F_i Q F_i^T$$

where $Q = \begin{bmatrix} Q_1 & \cdots & 0 \\ \vdots & \ddots & \vdots \\ 0 & \cdots & Q_n \end{bmatrix}$.

**2) ESKF update**

Once the propagation process is completed, the update process is triggered. Since the range measurements come at a higher frequency than wheel encoders, we put them in a weighted smoothing queue. The update process then utilizes the smoothed and the newest range measurements, to construct the Jacobians **H** of observation function $h()$ with respect to the vehicles' poses as follows.

Two kinds of range measurements exist. One is the range measurement between two dynamic vehicles, which gives rows of the Jacobians in the form as: $H_{ij} = [0_{1\times 3(i-1)} \quad e_{ij} \quad 0_{1\times 3(j-i-1)} \quad -e_{ij} \quad 0_{1\times 3(n-j)}](i < j)$, where $e_{ij}$ is the Jacobians of $h(x_i, x_j)$ with respect to poses of vehicle $R_i$ and $R_j$, $e_{ij} = \begin{bmatrix} \frac{p_i^T - p_j^T}{\|p_i - p_j\|} & 0 \end{bmatrix}^T$. The other kind is the range measurement between a dynamic vehicle $R_i$ and a static vehicle $R_k$, which gives: $H_{ik} = [0_{1\times 3(i-1)} \quad e_{ik} \quad 0_{1\times 3(n-i-1)}]$. By stacking all $H_{ij}$ and $H_{ik}$, the Jacobians **H** of the entire system can then be constructed and the standard procedure of the ESKF update can hereafter be followed.

**SIMULATION RESULTS**

In order to validate the preceding theoretical analysis and demonstrate the localization accuracy improvement due to collaboration, several numerical simulations are conducted. The simulation scenario consists of five homogenous ground vehicles, among them three vehicles move in an area of size 12*12 square meters, and two vehicles are static. Each vehicle is equipped with wheel encoders, which provide noisy measurements of linear velocity and turn rate, with a standard deviation of 0.2 m/s and 0.1 rad/s. And the inter-vehicle range measurements are corrupted with Gaussian noise and the standard deviation is 0.1 m. For simplicity, we assume that all measurements occur at the same sampling point and are fed into our range-only collaborative localization algorithm. Note that we assume that the initial states of each vehicle are known, since they are not the focus of the simulation.

Fig. 3 shows the case where the system is not fully observable. The left figure shows the estimated trajectories of three dynamic vehicles obtained from a typical simulation, in which case the range measurements between dynamic vehicles and static vehicles are not used. The right figure shows the estimated trajectories when the range measurements from the dynamic vehicles and a static vehicle are used. Clearly, in both case the estimated trajectories drift with time and the error is not bounded.

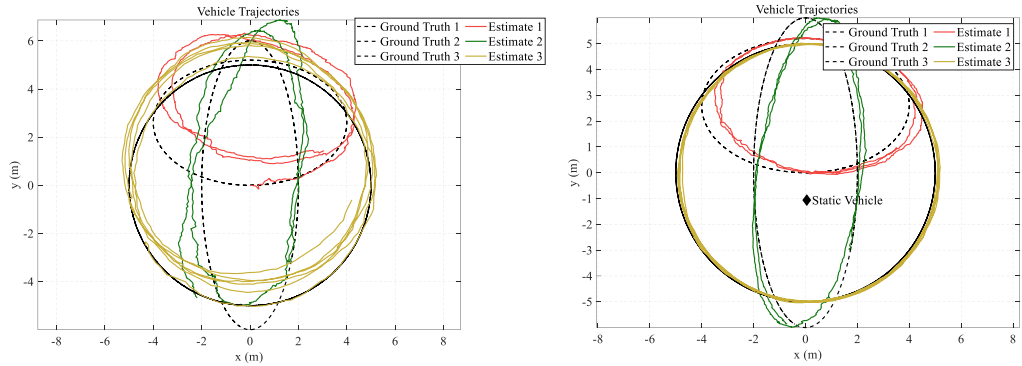

Fig.3. Estimated trajectories of collaborative localization when the system is not fully observable.

Fig. 4 illustrate the estimated trajectories and the root-mean-square error (RMSE), in which case all the range measurements are utilized. It can be seen that the estimated trajectories nearly coincident with the ground truth. And the collaborative localization errors for all the three vehicles are bounded, and quite below the error of integrated odometry results from pure motion measurements. This is attributed to the condition that at least two vehicles are required to be static to provide full observability of the range-only collaborative localization system. And the significance of collaboration is quite evident when compared to the odometry results.

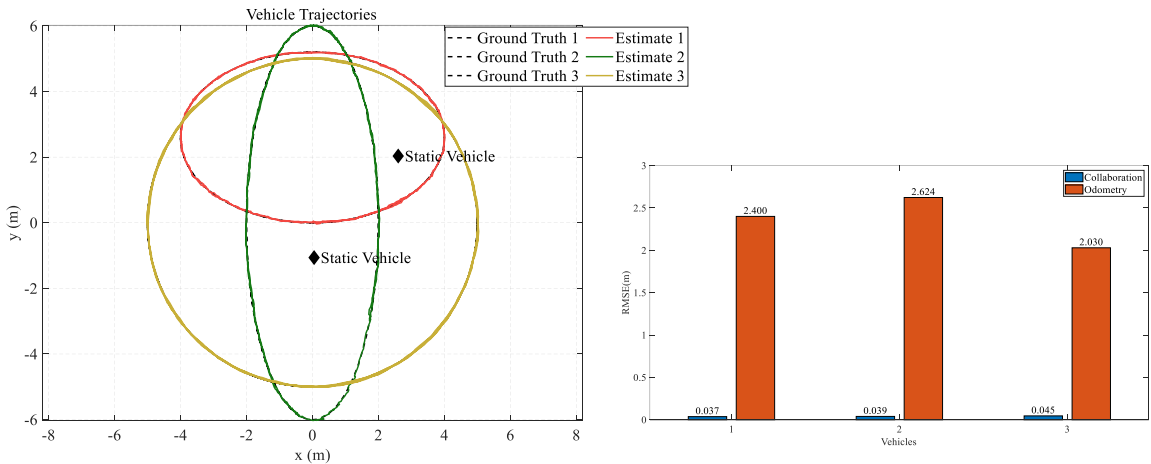

Fig.4. Left: Estimated trajectories of collaborative localization when the system is fully observable, in which case two vehicles remain static; Right: Comparison between the collaborative localization and the integrated odometry results from pure motion measurements.

**EXPERIMENTAL RESULTS**

In what follows, we carry out some real-world experiments to further evaluate the performance of our proposed range-only collaborative localization. We first consider a team of three connected ground vehicles. The experimental configuration is shown in Fig 5. Without loss of generality, we select the red vehicle ($R_1$) and the yellow vehicle ($R_3$) as static vehicles, the green ($R_4$) is manually controlled to randomly move in a rectangular area of 4m × 5m while avoiding collision with its teammates as well as the obstacles in the area. Each vehicle is equipped with a UWB transceiver and wheel encoders. For performance comparison, the vehicle is equipped with an additional Lidar. An onboard Lidar SLAM algorithm is operated to provide a reference ground truth by finding the correspondence to a pre-scanned environment Lidar map. The Lidar SLAM outputs the vehicle's pose at a frequency of 2Hz, with an approximate standard deviation of 0.1m for the position. We note that the Lidar SLAM results are time stamped by our UWB transceiver sending one-pulse-per-second (1PPS) timing packets to the vehicle, thus are time synchronized to the collaborative localization results.

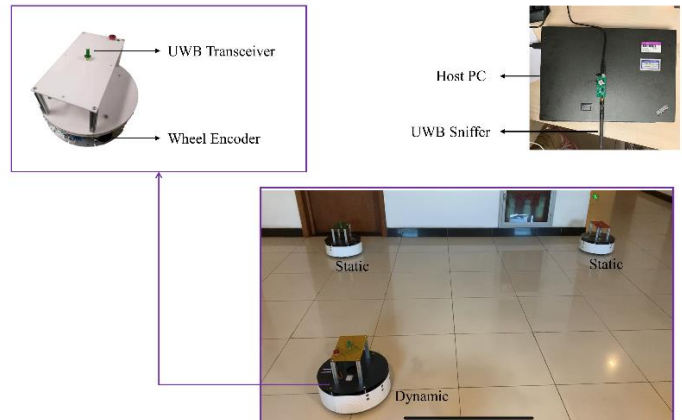

Fig.5. The experimental configuration with three UWB connected ground vehicles with two vehicles remain static (red and green) and one dynamic vehicle (yellow). Each vehicle is equipped with a UWB transceiver for both ranging and communication. The information is collected using a UWB sniffer on the host PC, which runs a centralized collaborative localization estimator.

The centralized range-only collaborative localization estimator is implemented using C++. It outputs the real-time trajectory of the dynamic vehicle $R_4$ as shown in Fig.6. The absolute position error is evaluated by comparing the estimation results of our algorithm and the Lidar SLAM at the same time. Fig. 7 shows the absolute pose error for $R_4$. The RMSE evaluated by absolute position error (APE) is 0.139m and the RMSE evaluated by absolute heading error is 0.051 rad, which is an impressive result in an approximate 5min run in total. The experiment shows that our range-only collaborative localization system is capable of real-time estimating the states of ground vehicles with an accuracy of decimeter, thus proving its usability in real-world. This also agrees with the observability analysis in the preceding section, since the results are with bounded error.

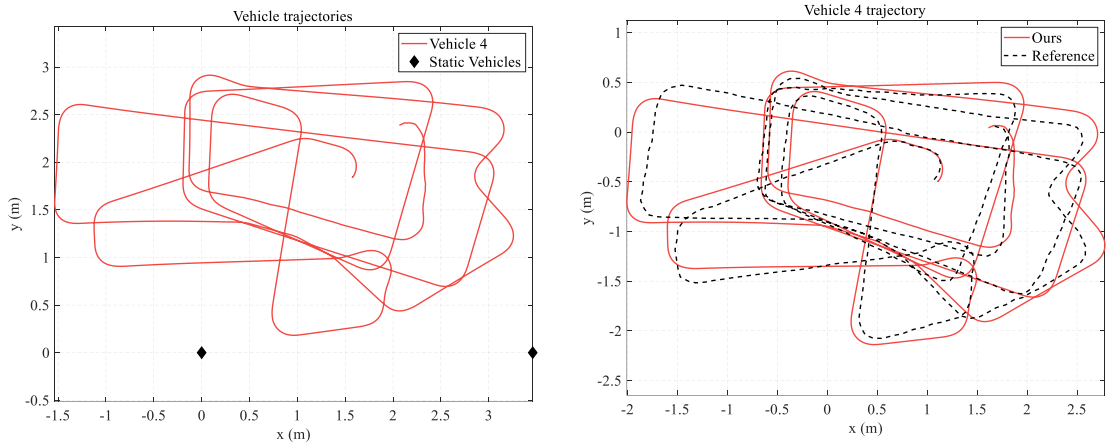

Fig.6. Left: the estimated trajectory of the dynamic vehicle and the positions of the static vehicles; Right: Our estimated results of $R_4$ compared with ground truth reference.

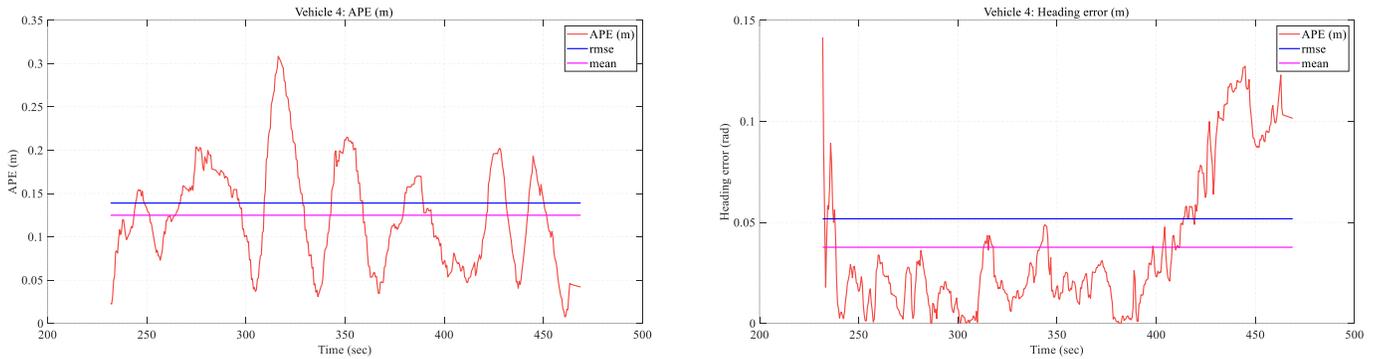

Fig.7. Absolute pose error of the dynamic vehicle. The two plots are absolute errors in position and heading.

To show the scalability of our system, we further perform an experiment with a team of five connected vehicles, in which case two vehicles remain static and three vehicles are moving in the area. The estimated trajectories are shown in Fig. 8. Since the Lidar SLAM is not compatible with such a dynamic environment, we do not have ground truth of the vehicles. Therefore, we control the three vehicles moving in a circular pattern during the experiment. As we can see, the estimated trajectories of three dynamic vehicles are apparently circular at the end, which provides strong qualitative evidence of our algorithm's performance in terms of accuracy.

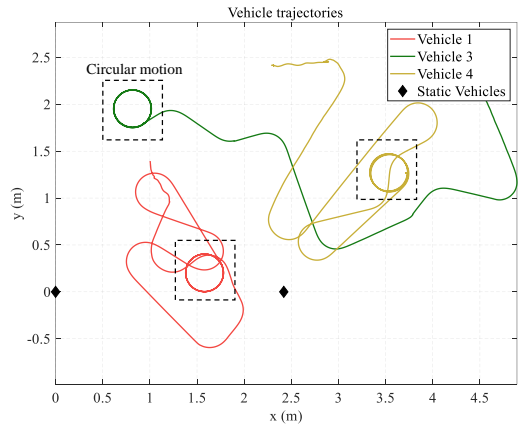

Fig.8. The estimated trajectories of three dynamic vehicles. They end up moving with a circular motion, which is verified by our collaborative localization results.

## CONCLUSION

In this paper, we propose a range-only collaborative localization for a team of UWB connected ground vehicles. We first analyze the observability of the system and show that to achieve full observability, two vehicles are required to remain static. We then implement our system utilizing a UWB network that provides both ranging and communication functionality. A centralized collaborative localization estimator is implemented with an initialization procedure. Simulation results have validated the observability conditions, and experimental results have proven the performance of our system in terms of accuracy and scalability.

As part of future work, we will conduct a distributed implementation of our system and extend our approach to the case of 3D localization.

## ACKNOWLEDGEMENT
This work is supported by Aerospace Science and Technology Foundation Grant.

**REFERENCES**

1. Michael N, Shen S, Mohta K, Mulgaonkar Y, Kumar V, Nagatani K, Okada Y, Kiribayashi S, Otake K, Yoshida K, and Ohno K, "Collaborative mapping of an earthquake-damaged building via ground and aerial robots," *Journal of Field Robotics*, Vol. 29, No. 5, Sep 2012, pp. 832-41.
2. Sun Z, Wang P, Vuran MC, Al-Rodhaan MA, Al-Dhelaan AM, and Akyildiz IF, "BorderSense: Border patrol through advanced wireless sensor networks," *Ad Hoc Networks*, Vol. 9, No. 3, May 2011, pp. 468-477.
3. Sand S, Zhang S, Mühlegg M, Falconi G, Zhu C, Krüger T, and Nowak S. "Swarm exploration and navigation on mars," *2013 International Conference on Localization and GNSS (ICL-GNSS), Jun 2013*, pp. 1-6.
4. Vásárhelyi G, Virágh C, Somorjai G, Nepusz T, Eiben AE, and Vicsek T. "Optimized flocking of autonomous drones in confined environments," *Science Robotics*, Vol. 3, No. 20, Jul 2018, eaat3536.
5. Preiss JA, Honig W, Sukhatme GS, and Ayanian N, "Crazyswarm: A large nano-quadcopter swarm," *2017 IEEE International Conference on Robotics and Automation (ICRA), May 2017*, pp. 3299-3304.
6. Jia, M, Zhao S, Dong D, Cui X, Lu M, "Positioning Algorithm Adaptation of an Indoor Navigation System for Virtual Reality Game Applications," *Proceedings of the 29th International Technical Meeting of the Satellite Division of The Institute of Navigation (ION GNSS+ 2016), Portland, Oregon, September 2016,* pp. 1824-1830.
7. Durrant-Whyte H, and Bailey T, "Simultaneous localization and mapping: part I," *IEEE robotics & automation magazine*, Vol. 13, No. 2, Jun 2006, pp. 99-110.
8. Li W, Cui X, Xu X, Lu M, "An Improved Ambiguity Resolution Algorithm Based on Particle Filter for INS/RTK Integration in Urban Environments," *Proceedings of the 31st International Technical Meeting of the Satellite Division of The Institute of Navigation (ION GNSS+ 2018), Miami, Florida, September 2018*, pp. 3122-3135.
9. Qin T, Li P, Shen S, "Vins-mono: A robust and versatile monocular visual-inertial state estimator," *IEEE Transactions on Robotics*, Vol. 34, No. 4, Aug 2018, pp. 1004-1020.
10. Roumeliotis SI and Bekey GA, "Distributed multirobot localization," *IEEE transactions on robotics and automation*, Vol. 18, No. 5, Oct 2002, pp. 781-795.
11. Huang GP, Trawny N, Mourikis AI and Roumeliotis SI, "Observability-based consistent EKF estimators for multi-robot cooperative localization," *Autonomous Robots*, Vol. 30, No. 1, Jan 2011, pp. 99-122.
12. Krajník T, Nitsche M, Faigl J, Vaněk P, Saska M, Přeučil L, Duckett T and Mejail M, "A practical multirobot localization system," *Journal of Intelligent & Robotic Systems*, Vol. 76, No. 3-4, Dec 2014, pp. 539-62.
13. Gross JN, Gu Y and Rhudy MB. "Robust UAV relative navigation with DGPS, INS, and peer-to-peer radio ranging," *IEEE Transactions on Automation Science and Engineering*, Vol. 12, No. 3, Jul 2015, pp. 935-944.
14. Nilsson JO, Zachariah D, Skog I and Händel P, "Cooperative localization by dual foot-mounted inertial sensors and inter-agent ranging," *EURASIP Journal on Advances in Signal Processing*, Vol. 2013, No. 1, Dec 2013, pp. 164.
15. Lee JY and Scholtz RA. "Ranging in a dense multipath environment using an UWB radio link," *IEEE Journal on Selected Areas in Communications*, Vol. 20, No. 9, Dec 2002, pp. 1677-1683.
16. Huang GP, Trawny N, Mourikis AI and Roumeliotis SI, "Observability-based consistent EKF estimators for multi-robot cooperative localization," *Autonomous Robots*, Vol. 30, No. 1, Jan 2011, pp. 99-122.
17. Martinelli A, Pont F and Siegwart R, "Multi-robot localization using relative observations," *Proceedings of the 2005 IEEE international conference on robotics and automation, Apr 2005,* pp. 2797-2802.
18. Zhou XS and Roumeliotis SI, "Robot-to-robot relative pose estimation from range measurements," *IEEE Transactions on Robotics,* Vol. 24, No. 6, Dec 2008, pp. 1379-1393.
19. Sharma R, Beard RW, Taylor CN and Quebe S, "Graph-based observability analysis of bearing-only cooperative localization", *IEEE Transactions on Robotics,* Vol. 28, No. 2, Dec 2011, pp. 522-529.
20. Jiang Y and Leung VC, "An asymmetric double sided two-way ranging for crystal offset," *2007 International Symposium on Signals, Systems and Electronics, Jul 2007,* pp. 525-528.
21. Hermann R and Krener A, "Nonlinear controllability and observability," *IEEE Transactions on automatic control*, Vol. 22, No. 5, Oct 1977, pp. 728-740.
22. Piziak R and Odell L, "Full rank factorization of matrices", *Mathematics Magazine,* Vol. 72, No. 3, Jun 1999, pp. 193-201.
23. Dokmanic I, Parhizkar R, Ranieri J and Vetterli M, "Euclidean distance matrices: essential theory, algorithms, and applications," *IEEE Signal Processing Magazine*, Vol. 32, No. 6, Nov 2015, pp. 12-30.
24. Sola J, "Quaternion kinematics for the error-state Kalman filter," arXiv preprint arXiv:1711.02508. Nov 2017.